\documentclass{llncs}
\usepackage{makeidx}  % allows for indexgeneration
\usepackage{llncsdoc}
\usepackage{amsmath,graphicx}
\usepackage{amssymb}
\usepackage{algorithm}
\usepackage[noend]{algpseudocode}
\usepackage{multirow}
\usepackage{array}
\usepackage[skip=3pt]{caption}
\usepackage{verbatim}
\usepackage{tabularx}
\usepackage{todonotes}

\begin{document}

%%%%%%%%% TITLE 
\title{Deep Learning using K-space Based Data Augmentation for Automated Cardiac MR Motion Artefact Detection}

\newcommand*\samethanks[1][\value{footnote}]{\footnotemark[#1]}
\newcommand{\rowstyle}[1]{\gdef\currentrowstyle{#1}%
  #1\ignorespaces
}

\author{Ilkay Oksuz$^{1}$,  
Bram Ruijsink$^{1,2}$,  
Esther Puyol-Ant\'on$^{1}$,  
Aurelien Bustin $^{1}$,
Gastao Cruz$^{1}$,  
Claudia Prieto  $^{1}$,  
Daniel Rueckert $^{3}$,
Julia A. Schnabel$^{1}$,  
Andrew P. King$^{1}$ }

\institute{$^{1}$School of Biomedical Engineering \& Imaging Sciences , King\rq{}s College London, UK \\ $^{2}$ Guy's and St Thomas' Hospital NHS Foundation Trust, London, UK \\ $^{3}$ Biomedical Image Analysis Group, Imperial College London, UK}

\maketitle

%%%%%%%%% ABSTRACT
\begin{abstract}

Quality assessment of medical images is essential for complete automation of image processing pipelines. For large population studies such as the UK Biobank, artefacts such as those caused by heart motion are problematic and manual identification is tedious and time-consuming. Therefore, there is an urgent need for automatic image quality assessment techniques. In this paper, we propose a method to automatically detect the presence of motion-related artefacts in cardiac magnetic resonance (CMR) images. As this is a highly imbalanced classification problem (due to the high number of good quality images compared to the low number of images with motion artefacts), we propose a novel k-space based training data augmentation approach in order to address this problem. Our method is based on 3D spatio-temporal Convolutional Neural Networks, and is able to detect 2D+time  short axis images with motion artefacts in less than 1ms.  We test our algorithm on a subset of the UK Biobank dataset consisting of 3465 CMR images  and achieve not only high accuracy in detection of motion artefacts, but also high precision and recall. We compare our approach to a range of state-of-the-art quality assessment methods. 

\end{abstract}


%%%%% INTRODUCTION %%%%%

\section{Introduction}  \label{sec:intro}

High diagnostic accuracy of image analysis pipelines requires high quality medical images.  Misleading conclusions can be drawn when the original data are of low quality, in particular for cardiac magnetic resonance (CMR) imaging. CMR images can contain a range of image artefacts \cite{Ferreira2013}, and assessing the quality of images acquired on MR scanners is a challenging problem. Traditionally, images are visually inspected by one or more experts, and those showing an insufficient level of quality are excluded from further analysis. However, visual assessment is time consuming and prone to variability due to inter-rater and intra-rater differences.

\begin{figure}[htb]

\begin{minipage}[b]{.31\linewidth}
  \centering
  \centerline{\includegraphics[width=4.0cm]{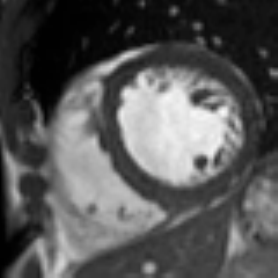}}
%  \vspace{1.5cm}
  \centerline{(a) Good quality image}\medskip
\end{minipage}
\hfill
\begin{minipage}[b]{0.31\linewidth}
  \centering
  \centerline{\includegraphics[width=4.0cm]{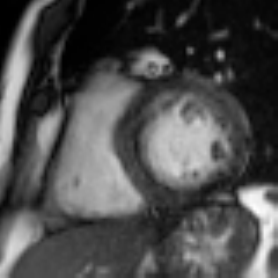}}
%  \vspace{1.5cm}
  \centerline{(b) Motion artefact image}\medskip
\end{minipage}
\hfill
\begin{minipage}[b]{0.31\linewidth}
  \centering
  \centerline{\includegraphics[width=4.0cm]{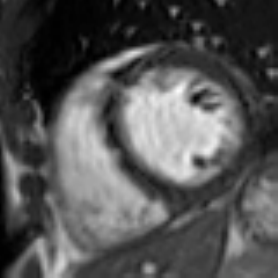}}
%  \vspace{1.5cm}
  \centerline{ (c) K-space corrupted image} \medskip
\end{minipage}
\hfill

\caption{Examples of a good quality CINE CMR image (a), an image with blurring motion artefacts (b), and a k-space corrupted image (c). The k-space corruption process is able to simulate realistic motion-related artefacts. (Please see videos in supplementary material.)}
\label{fig:Motivation}
\end{figure}

The UK Biobank is a large-scale study with all data accessible to researchers worldwide, and will eventually consist of CMR images from 100,000 subjects \cite{Petersen2016}. To maximize the research value of this and other similar datasets, automatic quality assessment tools are essential. One specific challenge in CMR is motion-related artefacts such as mistriggering, arrhythmia and breathing artefacts. These can result in temporal and/or spatial blurring of the images, which makes subsequent processing difficult \cite{Ferreira2013}.  For example, this type of artefact can lead to erroneous quantification of left ventricular (LV) volumes, which is an important indicator for cardiac functional assessment. Examples of a good quality image and an image with blurring motion artefacts are shown in Fig.\ref{fig:Motivation}a-b for a short-axis view CINE CMR scan.
 
%%%%% Background %%%%%
\section{Background} \label{sec:background}

 Convolutional Neural Networks (CNNs) have been utilized for image quality assessment in the computer vision literature with considerable success \cite{Kang2014}, especially for image compression detection. This success has motivated the medical image analysis community to utilize CNNs on multiple image quality assessment challenges such as fetal ultrasound \cite{Wu2017} and echocardiography \cite{Abdi2017a}. CNNs have also been utilized for real-time scan plane detection for ultrasound with high accuracy \cite{Baumgartner2016}. Relevant work in motion artefact detection includes Meding et al. \cite{Meding2017}, who proposed a technique to detect motion artefacts caused by patient movement for brain MR using a simple CNN architecture. Their algorithm used training data from fruit images and artificial motion data. Also, Kustner et al. \cite{Kustner2017} proposed a patch-based motion artefact detection method for brain and abdomen MR images, but they made their tests on a small dataset of 16 MR images with significant motion artefacts. Motion artefacts in CMR imaging are largely caused by ECG mistriggering and pose a significantly different problem.  
 
 In the context of CMR, recent work on image quality assessment has targeted the detection of missing apical and basal slices \cite{Zhang2017}. Missing slices adversely affect the accurate calculation of the LV volume and hence the derivation of cardiac functional metrics such as ejection fraction. Tarroni et al. \cite{Tarroni2017} addressed the same problem using multiple view CMR images and decision trees for landmark localization. One other image quality issue is off-axis acquisition of 4-chamber view CMR, which was detected with a simple CNN in \cite{Oksuz2018}. CMR image quality was also linked with automatic quality control of image segmentation in \cite{Robinson2017}.  In the context of  detection of CMR motion artefacts,  Lorch et al. \cite{Lorch2017} investigated synthetic motion artefacts. In their work, they used histogram, box, line and texture features to train a random forest algorithm to detect motion artefacts for different artefact levels. However, their algorithm was tested only on artificially corrupted synthetic data and aimed only at detecting breathing artefacts.

In this paper, we aim to accurately detect motion artefacts in large CMR datasets. We use a CNN for the detection of such cases and evaluate our method on a dataset of 3465 2D+time CMR sequences from the UK Biobank. There are two major contributions of this work. First, we address the problem of fully automatic motion artefact detection in a real large-scale CMR dataset for the first time. Second, we introduce a realistic k-space based data augmentation step to address the class imbalance problem.  We generate  artefacts from good quality images using a k-space corruption scheme, which results in highly realistic artefacts as visualized in Fig.\ref{fig:Motivation}c.

%%%%% Methods %%%%%

\section{Methods}
\label{sec:method}

The proposed framework of using a spatio-temporal CNN for motion artefact detection consists of two stages; 1) image preprocessing consisting of normalization and region-of-interest (ROI) extraction; 2) CNN image classification of motion artefacts and good quality images.

\subsection{Image Preprocessing}
Given a 2D CINE CMR sequence of images we first normalize the pixel values between 0 and 1. Since the image dimensions vary from subject to subject, instead of image resizing we propose to use a motion information based ROI extraction to $80\times 80$ pixels. Avoiding image resizing  is of particular importance for image quality assessment, because resizing can influence the image quality significantly. The ROI is determined by using an unsupervised technique based on motion as proposed in \cite{Korshunova2016}. Briefly, the ROI is determined by performing a Fourier analysis of the sequences of images, followed by a circular Hough transform to highlight the center of periodically moving structures.

\subsection{Network Architecture}

\begin{figure}[htb]
  \centering
  \centerline{\includegraphics[width=\linewidth]{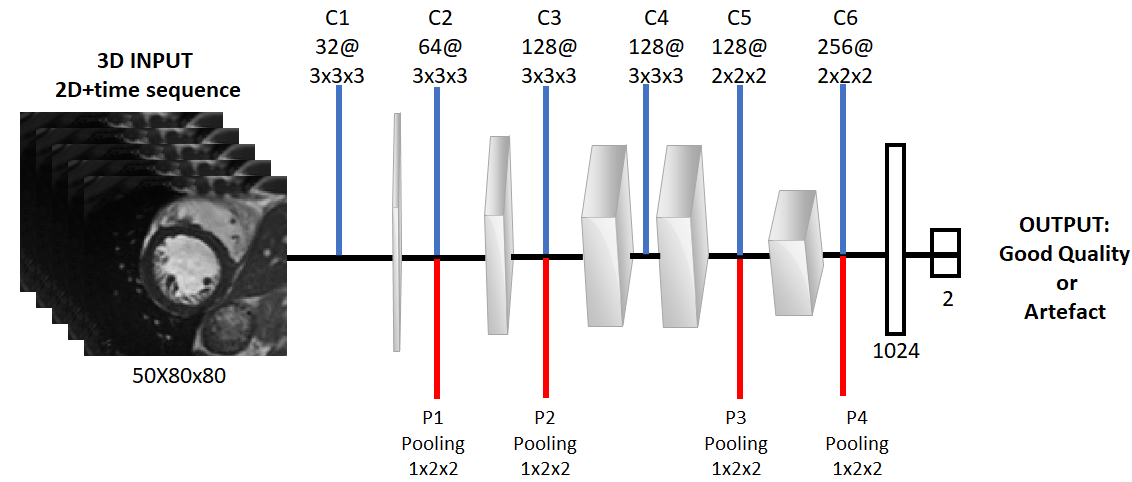}}
\caption{The CNN architecture for motion artefact detection. }
\label{fig:Model}
\end{figure}

The proposed CNN consists of eight layers as visualized in Fig. \ref{fig:Model}. The architecture of our network follows a similar architecture to \cite{Tran2017}, which was originally developed for video classification using a spatio-temporal 3D CNN. In our case we use the third dimension as the time component and use 2D+time mid-ventricular sequences for classification. The input is an intensity normalized $80 \times 80$ cropped CMR image with 50 time frames. The network has 6 convolutional layers and 4 pooling layers, 2 fully-connected layers and a softmax loss layer to predict motion artefact labels.  After each convolutional layer a ReLU activation is used. We then apply pooling on each feature map to reduce the filter responses to a lower dimension. We  apply dropout with a probability of 0.5 at all convolutional layers and after the first fully connected layer to enforce regularization. All of these convolutional layers are applied with appropriate padding (both spatial and temporal) and stride 1, thus there is no change in terms of size from the input to the output of these convolutional layers.

\subsection{K-space Corruption for Data Augmentation}
\label{sect:synth}
\begin{figure}[htb]
  \centering
  \centerline{\includegraphics[width=\linewidth]{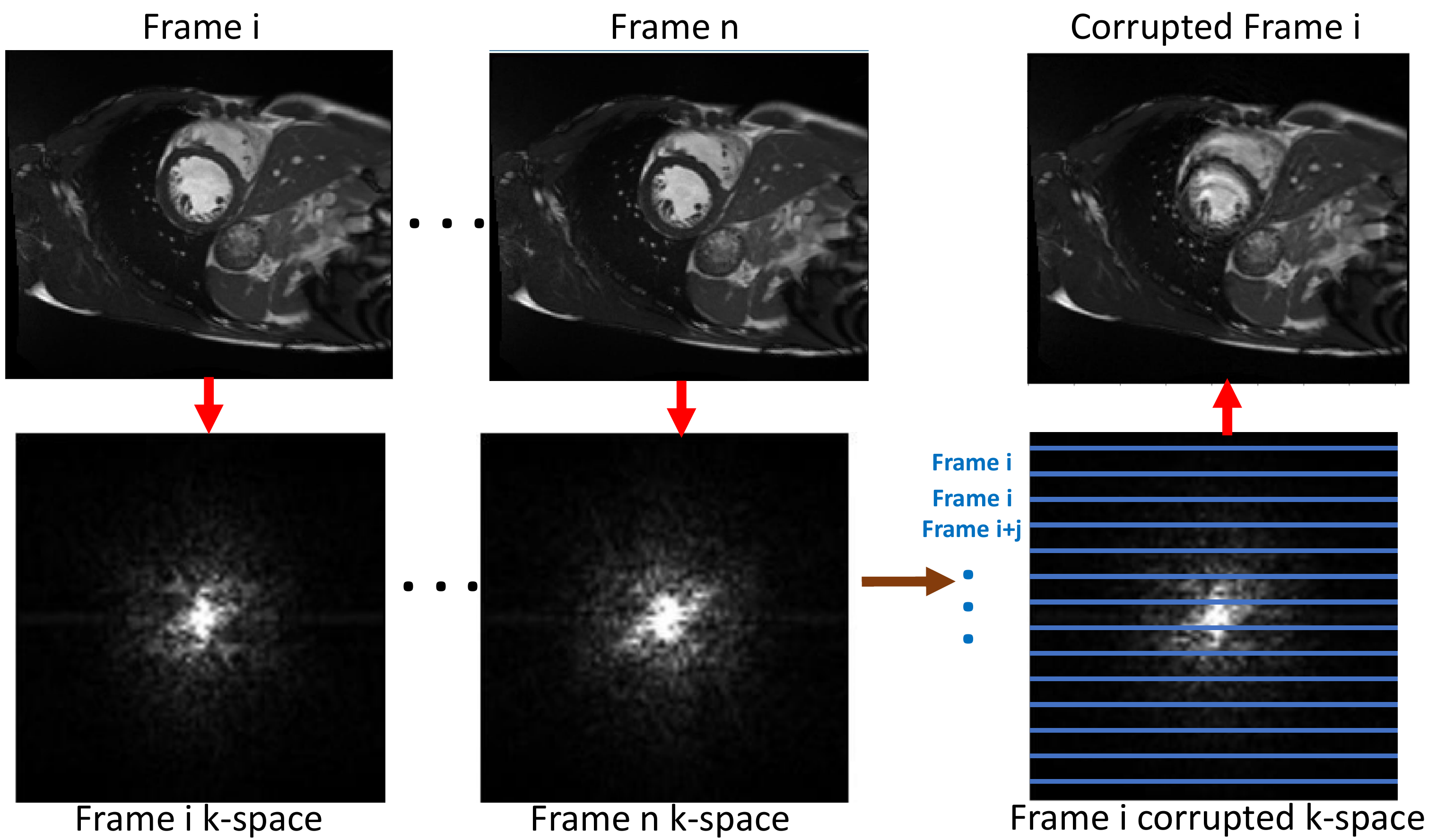}}
\caption{K-space corruption for motion artefact generation in k-space. We use the Fourier transformation of each frame to generate the k-space of each image and replace k-space lines with lines from different temporal frames to generate corruptions. }
\label{fig:Kspace}
\end{figure}

We generate k-space corrupted data in order to increase the amount of motion artefact data for our under-represented low quality image class. The UK Biobank dataset that we use was acquired using Cartesian sampling and we follow a Cartesian k-space corruption strategy to generate synthetic but realistic motion artefacts. We first transform each 2D short axis sequence to the Fourier domain and change 1 in $z$ Cartesian sampling lines to the corresponding lines from other cardiac phases to mimic motion artefacts. We set $z=3$ for generating realistic corruptions. In Fig. \ref{fig:Kspace} we show an example of the generation of a corrupted frame $i$ from the original frame $i$ using information from the k-space data of other temporal frames. We add a random frame offset $j$ when replacing the lines. In this way the original good quality images from the training set are used to increase the total number of  low quality images in the training set. This is a realistic approach as the motion artefacts that occur from mistriggering arise from similar misallocations of k-space lines.

\subsection{Data Augmentation}
\label{sect:aug}

Data augmentation was applied to achieve a more generalized model.  Translational invariance is ensured with random shifts in both the horizontal and vertical directions in the range of [W/5, H/5], where W and H represent the width and height of the image respectively (i.e. W=H=80 pixels in our case). Rotations are not used due to their influence on image quality caused by the necessary interpolation. Balanced classes are ensured using  augmentation with the k-space corrupted data (see Section \ref{sect:synth}) for the under-represented motion artefact class. Note that none of the augmented data were used for testing. They were only used for increasing the total number of training images.

\subsection{Implementation Details}

The training of a CNN can be viewed as a combination of two components, a loss function or training objective, and an optimization algorithm that minimizes this function. In this study, we use the Adadelta optimizer  to minimize the binary cross entropy. The cross entropy represents the dissimilarity of the approximated output distribution  from the true distribution of labels after a softmax layer.  The training ends when the network does not significantly improve its performance on the validation set for a predefined number of epochs (50). An improvement is considered sufficient if the relative increase in performance is at least 0.5\%. 

During training, a batch-size of 50 2D+time sequences was used. The  momentum of the optimizer was set to 0.90 and the learning rate was 0.0001. The parameters of the convolutional and fully-connected layers were initialized randomly from a zero-mean Gaussian distribution. In each trial, training was continued until the network converged. Convergence was defined as a state in which no substantial progress was observed in the training loss. We used the Keras Framework with Tensorflow backend for implementation and training the network took around 4 hours on a NVIDIA Quadro 6000P GPU. Classification of a single image sequence took less than 1ms.

\section{Experimental Results}
\label{sec:results}

We evaluated our algorithm on a subset of the UK Biobank dataset consisting of 3360 good quality acquisitions and 105 short axis acquisitions containing arrythmia, mistriggering and breathing artefacts. This subset was chosen to be free of other types of image quality issues such as missing axial slices and was visually verified by an expert cardiologist. The details of the acquisition protocol of the UK Biobank dataset can be found in \cite{Petersen2016}. 50 temporal frames of each subject at mid-ventricular level were used to detect the motion artefacts. 

We tested four different training configurations of our proposed CNN technique to evaluate its performance in more detail: (1) training using only real data without any data augmentation, (2) training using data augmentation with translations only, (3) training using data augmentation with k-space corrupted data only, (4) training using data augmentation with both translations and k-space corrupted data. In each setup, the real motion artefact data is used.

We compared our algorithm with a range of alternative classification techniques: K-nearest neighbours, Support Vector Machines (SVMs), Decision Trees, Random Forests, Adaboost  and  Naive Bayesian. The inputs to all algorithms were the cropped intensity-normalized data as described in Section 3.1. We optimized the parameters of each algorithm using a grid search. We also tested two techniques developed for image quality assessment in the computer vision literature: the NIQE metric \cite{Mittal2013} is based on natural scene statistics and is trained using a set of good quality CMR images to establish a baseline for good image quality; and the Variance of Laplacians is a moving filter that has been used to detect the blur level of an image. For both of these techniques we used a SVM for classification of the final scores.

We used a 10-fold stratified cross validation strategy to test all algorithms, in which each image appears once in the test set over all folds. Due to the high class imbalance, over 0.9 accuracy is achieved by most of the techniques in Table \ref{table:1}. We do not rely only on accuracy in our results due to the bias introduced by the imbalanced classes. The interesting comparison for the methods lies in the recall numbers, which quantify the capability of the methods to identify images with artefacts. The results show that the CNN-based technique is capable of identifying the presence of motion artefacts with high recall compared to other techniques. Note in particular that introducing the data with k-space corruption increases the recall number from 0.466 to 0.642. Additional data augmentation improves the recall further to 0.652.  All state-of-the-art classification methods reach high accuracy, but Adaboost is capable to reach comparable recall to CNNs without data augmentation. However, the precision of Adaboost is lower compared to the CNN results. Moreover, the F1-score illustrates the improvement of performance using CNN-based methods compared to state-of-the-art machine learning approaches. The F1-score reaches 0.704 with k-space corrupted data and achieves 0.722 with additional data augmentation. 

\begin{table} 

\centering
\caption{Mean accuracy, precision, recall and F1 score results of image classification for motion artefacts. 10-fold cross validation is used and each image is labelled once over all folds.}
\begin{tabular}{lcccc}
\hline
Methods   & Accuracy & Precision & Recall & F1-score \\
\hline 

K-Nearest Neighbours  & $0.952$  & $0.074$  & $0.268$  & $0.116$    \\
Linear SVM    & $ 0.968 $  & $0.721$  & $0.385$  & $0.502$  \\
Decision Tree   &   $ 0.951 $  & $0.250$  & $0.385$  & $0.303$  \\
Random Forests   & $0.958$    & $0.320$  & $0.315$  & $0.317$     \\
Adaboost   & $0.960 $  & $0.230$  & $0.567$  & $0.327$       \\
Naive Bayesian   & $0.801 $  & $0.527$  & $0.183$  & $0.111$       \\
Variance of Laplacian  & $0.958$  & $0.113$  & $0.161$  & $0.133$    \\
NIQE \cite{Mittal2013}  & $0.958$  & $0.210$  & $0.248$  & $0.227$    \\
\hline
\textbf{CNN-no augmentation} \cite{Tran2017}    & $0.968$  & $0.700$  & $0.466$  & $0.560$      \\
\textbf{CNN-translational augmentation}    & $0.974$  & $0.750$  & $0.600$  & $0.667$     \\
\textbf{CNN-k-space augmentation}      & $0.977$  & $0.779$ & $0.642$ & $0.704$   \\
\textbf{CNN with  k-space+translational augmentation}    & \textbf{0.982}  &  \textbf{0.809} &  \textbf{0.652} &  \textbf{0.722} \\
\hline
\end{tabular}
\label{table:1}
\end{table}

%\vspace{-0.2cm}
\section{Discussion and Conclusion}
\label{sec:discussion}
%\vspace{-0.2cm}

In this paper, we have proposed a CNN-based technique for identifying motion-related artefacts in a large 2D CINE CMR dataset with high accuracy and recall. We have addressed the issue of high class imbalance by proposing the use of a k-space corruption data generation scheme for data augmentation, which clearly outperforms the basic data augmentation due to its capability to generate realistic artefact images. We have shown that a 3D CNN based neural network architecture developed for video classification  is capable of classifying motion artefacts, outperforming other state-of-the-art techniques.  To the best knowledge of the authors, this is the first paper to have tackled this clinical problem for a real CMR dataset. Our work brings fully automated evaluation of ventricular function from CMR imaging a step closer to clinically acceptable standards, enabling analysis of large imaging datasets such as the UK Biobank.

In future work, we plan to validate our method on the entire UK Biobank cohort, which is eventually expected to be 100,000 CMR images. Moreover, our current technique focuses only on mid-ventricular slices and in future we will use other slices in order to perform a more thorough image quality assessment.

\textbf{Acknowledgments}\\
%\begin{center}
This work was supported by an EPSRC programme Grant (EP/P001009/1) and the Wellcome EPSRC Centre for Medical Engineering at the School of Biomedical Engineering and Imaging Sciences, King’s College London (WT 203148/Z/16/Z). This research has been conducted using the UK Biobank Resource under Application Number 17806. The GPU used in this research was generously donated by the NVIDIA Corporation.
%\end{center}

\end{document}